\documentclass[times,twocolumn,final,authoryear]{elsarticle}

\usepackage{ycviu}
\usepackage{framed,multirow}

\usepackage{amssymb}
\usepackage{amsmath,amsfonts}
\usepackage{latexsym}

\usepackage{url}
\usepackage{xcolor}
\usepackage{tikz}
\usepackage{array}
\usepackage{caption}
\usepackage{multirow}
\usepackage{booktabs}
\usepackage{pdfpages}
\usepackage{pgfplots}
\usepackage{caption}
\usepackage{subcaption}
\usepackage{soul,color}
\pgfplotsset{compat=1.17}

\definecolor{newcolor}{rgb}{.8,.349,.1}

\newcommand{\bone}{\mathbf{1}}

\newcommand{\cB}{\mathcal{B}}

\newcommand{\bC}{\mathbf{C}}

\newcommand{\cD}{\mathcal{D}}

\newcommand{\bD}{\mathbf{D}}

\newcommand{\bbR}{\mathbb{R}}

\newcommand{\bK}{\mathbf{K}}

\newcommand{\tikzcircle}[2][red,fill=red]{\tikz[baseline=-0.7ex]\draw[#1,radius=#2, ultra thick] (0,0) circle ;}

\newcolumntype{P}[1]{>{\centering\arraybackslash}p{#1}}

\definecolor{myred}{HTML}{e74c3c}
\definecolor{mygreen}{HTML}{2ecc71}
\definecolor{mybluelight}{HTML}{3498db}
\definecolor{myblue}{HTML}{2980b9}
\definecolor{greendarnertail}{HTML}{74b9ff}
\definecolor{mylightyellow}{HTML}{fffa65}
\definecolor{mylightred}{HTML}{fab1a0}
\definecolor{mylightgreen}{HTML}{55efc4}
\definecolor{mylightpink}{HTML}{ff9ff3}
\definecolor{mylightcyan}{HTML}{81ecec}
\definecolor{mylightpurple}{HTML}{a29bfe}

\journal{Computer Vision and Image Understanding}

\begin{document}

\ifpreprint
  \setcounter{page}{1}
\else
  \setcounter{page}{1}
\fi

\begin{frontmatter}

\title{Assessing Domain Gap for Continual Domain Adaptation in Object Detection}

\author[1]{Anh-Dzung Doan\corref{cor1}} 
\cortext[cor1]{Corresponding author}
\ead{dzung.doan@adelaide.edu.au}
\author[2]{Bach Long Nguyen}
\author[3]{Surabhi Gupta}
\author[1]{Ian Reid}
\author[2]{Markus Wagner}
\author[1]{Tat-Jun Chin}

\address[1]{Australian Institute for Machine Learning, The University of Adelaide, Adelaide, SA 5000, Australia}
\address[2]{Department of Data Science and Artificial Intelligence, Monash University, Melbourne, VIC 3800, Australia}
\address[3]{Safran Electronics \& Defense Australasia, Botany, NSW 2019, Australia}

\received{1 May 2013}
\finalform{10 May 2013}
\accepted{13 May 2013}
\availableonline{15 May 2013}
\communicated{S. Sarkar}

\begin{abstract}
To ensure reliable object detection in autonomous systems, the detector must be able to adapt to changes in appearance caused by environmental factors such as time of day, weather, and seasons. Continually adapting the detector to incorporate these changes is a promising solution, but it can be computationally costly. Our proposed approach is to selectively adapt the detector only when necessary, using new data that does not have the same distribution as the current training data. To this end, we investigate three popular metrics for domain gap evaluation and find that there is a correlation between the domain gap and detection accuracy. Therefore, we apply the domain gap as a criterion to decide when to adapt the detector. Our experiments show that our approach has the potential to improve the efficiency of the detector's operation in real-world scenarios, where environmental conditions change in a cyclical manner, without sacrificing the overall performance of the detector. Our code is publicly available~\url{https://github.com/dadung/DGE-CDA}.
\end{abstract}


\end{frontmatter}


\section{Introduction}
\label{sec:intro}

Object detection is the task of localising instances of objects of a certain category within images. It plays a key role in the scene understanding, which is a fundamental problem in various applications, such as, autonomous driving~\citep{arnold2019survey}, AR/VR~\citep{zhang2022sear}, and robotics~\citep{sachdeva2022autonomy}. A common approach to object detection is to train a deep neural network on a large dataset with ground truth bounding boxes and object categories. Current state-of-the-art methods include Faster-RCNN~\citep{fasterrcnn}, RetinaNet~\citep{retinanet}, YOLO~\citep{yolo}, etc.

In long-term operations, an object detector (OD) must be able to handle variations in appearance caused by natural factors such as time of day, seasons, and weather. These changes can cause the appearance of images to differ from the training data, leading to a decline in detection accuracy. To address this issue, recent works suggest continuously collecting new data and using it to update the perception systems~\citep{doan2019scalable,doan2020hm,churchill2013experience}. An ideal approach is to manually label all new data (target data) and combine it with the current training samples (source data) to create a new training database, which is then used to fine-tune the OD~\citep{chen2022update}. However, the significant manual annotation effort required for this approach makes it impractical.

To address the impracticality of manual annotation, researchers have focused on two main strategies. The first strategy is to use weakly supervised learning, which only requires labels indicating the presence of certain object categories in the image~\citep{bilen2016weakly,kantorov2016contextlocnet, song2014learning}. While this reduces annotation cost, it results in detectors with only about half the accuracy of those trained with fully supervised methods. The second strategy is active learning~\citep{roy2018deep, yuan2021multiple, yu2022consistency, choi2021active}, which selects the most informative samples from the target data for annotation. Typically, target images that cause the greatest uncertainty in the OD are chosen for annotation. However, as object detection involves both regression and classification, accurately characterising the uncertainty of predicted bounding boxes is a complex task.

Despite its potential, active learning still requires human effort in the adaptation process. To achieve a fully autonomous adaptation mechanism, unsupervised domain adaptation (UDA)~\citep{oza2021unsupervised} has gained popularity in the research community. In particular, adversarial feature learning~\citep{chen2018domain, chen2021scale,PASQUALINO2021104098} aims to minimise the domain gap between source and target data in the feature space by introducing an adversarial loss with respect to a domain discriminator. A gradient reversal layer~\citep{ganin2015unsupervised} is used to effectively incorporate adversarial learning into the backpropagation process. However, without proper tuning of the hyperparameters, this approach will likely struggle to achieve satisfactory results on challenging datasets.

Furthermore, image-to-image translation is a widely used approach in UDA for object detection~\citep{arruda2019cross,gao2021cyclegan,schutera2020night}. It uses the GAN principle~\citep{gan} to transform the appearance of images from the target domain to resemble the source domain. However, training a GAN-based model is a complex task due to its minimax optimisation problem. This can lead to various challenges such as non-convergence, mode collapse, and diminished gradient.

Another popular technique in UDA for object detection is self-training~\citep{roychowdhury2019automatic, khodabandeh2019robust, d2020one}. It involves using the detector trained on the source data to generate pseudo-labels for the target data. As the pseudo-labels are often noisy, heuristics are proposed to filter out potentially incorrect ones. However, this technique introduces a potential point of failure, as the heuristics may not be able to filter all noisy pseudo-labels, leading to errors that accumulate over time during long-term operations.

In general, existing works primarily focus on efficiently adapting ODs, assuming that the adaptation will always be carried out when the OD encounters a new target domain. However, in practical scenarios, changes in environmental conditions are often cyclical, such as different time of day, weather, and seasons. This means that the appearance of target images will likely resemble that of the source images after a period of time. For example, if the source images were captured in clear conditions from 10am to 3pm, target images captured in cloudy conditions at 1pm will likely be closer to the source images than target images captured in clear conditions at 11pm. This raises the question of whether it is necessary to always adapt the OD during its course of operation. The domain adaptation should only be performed if it can significantly improve the OD's accuracy or a new domain has a considerable negative impact on the detection performance. Therefore, selectively adapting the OD can save a significant amount of computational cost.

\vspace{0.3em}
\noindent\textbf{Contributions} \, Prior techniques mainly focus on ``how to efficiently adapt ODs?'' leaving the question ``when to adapt ODs?'' unaddressed. Therefore, in this paper, we will investigate the latter. Our hypothesis is that if the distribution of target data is similar to current training data, adapting ODs will not lead to significant improvements in accuracy while resulting in the unnecessary use of resources. To address this, we propose using domain gap as a criterion for deciding when adaptation is necessary in continually changing environmental conditions, based on our finding of a correlation between domain gap and detection accuracy. Our experiments demonstrate that this solution can save adaptation cost without sacrificing the overall performance of the OD.

\section{Related work}
\label{sec:related_work}
In order to achieve a robust object detector (OD) for long-term operations, it is vital to continually adapt the OD. There are three main approaches to do this in the literature: weakly supervised object detection (WSOD), active learning for object detection (ALOD), and unsupervised domain adaptation for object detection (UDAOD).

In WSOD, human effort is required to provide image-level labels, indicating which object categories are present in the image. Then, the problem is addressed through Multiple Instance Learning (MIL). The standard pipeline of WSOD consists of two phases~\citep{bilen2014weakly,song2014learning,cinbis2016weakly}. Firstly, a set of candidate bounding boxes (instances) likely containing objects are generated by object proposal methods (e.g., selective search~\citep{uijlings2013selective} or Edge Boxes~\citep{zitnick2014edge}). Secondly, the proposal classification phase alternates between two steps: \emph{i)} positive proposals with high confident scores are selected as pseudo instance-level labels and \emph{ii)} the OD (instances classifier) is trained under MIL framework. In this pipeline, \cite{bilen2016weakly} propose an end-to-end weakly
supervised deep detection network. The main idea is to estimate the image classification score as a linear combination of weighted proposal scores. Since then, a number of works have been proposed to improve the pipeline of~\cite{bilen2016weakly}, such as, multitasking with segmentation~\citep{gao2019c}, leveraging spatial information~\citep{kantorov2016contextlocnet}, click supervision~\citep{papadopoulos2017training}, and improving the proposals quality~\citep{cheng2020high}. Recently, \cite{inoue2018cross,hou2021informative,ouyang2021pseudo,xu2022h2fa} apply WSOD to the domain adaptation, where the performance of the OD in the target domain can be improved by using source data with full annotations.

Apart from WSOD, ALOD is another direction that aims to efficiently leverage human supervision in adapting ODs by exploiting model uncertainty to select the most informative images for human annotation. This helps to minimise the total annotation cost. Specifically, \cite{roy2018deep} consider the classification outputs of the detector to find images which the OD is uncertain most. Accordingly, \cite{roy2018deep} propose black-box and white-box methods, where the black-box method can be applied to a variety of network architectures while the white-box method is tailored for Single Shot Multibox Detector~\citep{liu2016ssd}. Concurrently, \cite{brust2018active} propose to aggregate confidence scores of bounding boxes to represent the model uncertainty. Then, \cite{yuan2021multiple} propose to re-weight bounding boxes to eliminate noisy ones before aggregation. Since previous works primarily focus on classification branch of ODs, \cite{choi2021active} employ mixture density networks to estimate the model uncertainty from both regression and classification branches.

In spite of a promising approach, ALOD still requires human effort in annotating data. To address this issue, UDAOD has become an increasingly popular approach in the community. Specifically, \cite{arruda2019cross} use CycleGAN~\citep{zhu2017unpaired} to translate image appearances from daytime to night to train an OD. However, CycleGAN can cause object of interests to be distorted, unrealistic, or disappeared, leading to noisy data for ODs. Therefore, detection-guided CycleGAN~\citep{gao2021cyclegan} adds a detection branch to guide the image generation process to ensure the quality of objects of interest in translated images. Another strategy is adversarial feature learning. In particular, domain adaptive Faster-RCNN~\citep{chen2018domain} minimises the domain discrepancy through aligning image and instance distributions. This is achieved via training domain classifiers in an adversarial manner. The idea of~\cite{chen2018domain} is then extended to a scale-aware method by~\cite{chen2021scale}, which aligns image distributions in different image scales through employing feature pyramid network~\citep{fpn}. A similar idea is also investigated in RetinaNet architecture~\citep{PASQUALINO2021104098}. Apart from adversarial feature learning, self-training has recently gained popularity in the community. Specifically, \cite{roychowdhury2019automatic} combine detection and tracking to generate pseudo-labels for target datasets. However, since pseudo-labels may be very noisy,  \cite{roychowdhury2019automatic} propose a label smoothing technique to mitigate this negative effect. Moreover, to more efficiently address the noise present in the pseudo-labels, \cite{khodabandeh2019robust} use an additional image classifier to improve the quality of pseudo-labels. Recently, \cite{li2022cross} employ the teacher-student model combined with self-training and adversarial feature learning to achieve a state-of-the-art result.

As alluded, continually adapting ODs is indeed an active research topic. However, existing studies primarily focus on developing efficient methods for adaptation and overlook the crucial question of when to adapt. In realistic scenarios, environmental conditions usually change in a cyclical manner. Hence, always adapting ODs is not only computationally expensive but also unlikely to produce significant accuracy improvement.

\section{Object detection architecture}
\begin{figure}[ht]\centering
	\centering
	
	    \subfloat[][]
		{
			\includegraphics[width=0.4\textwidth]{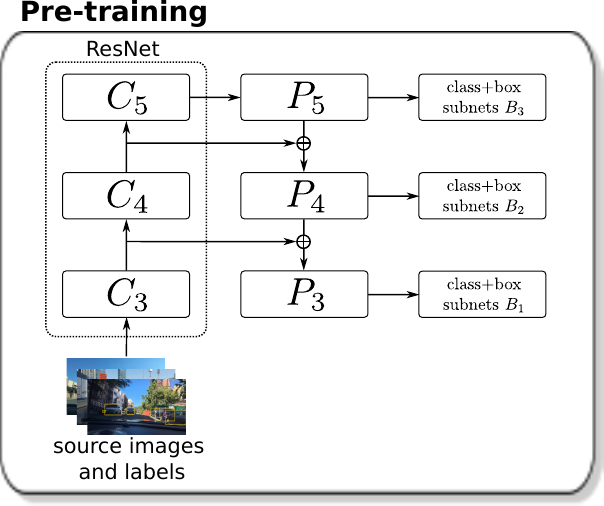}
			\label{fig:pipeline_pretraining}
		}
  
		\subfloat[][]
		{
			\includegraphics[width=0.4\textwidth]{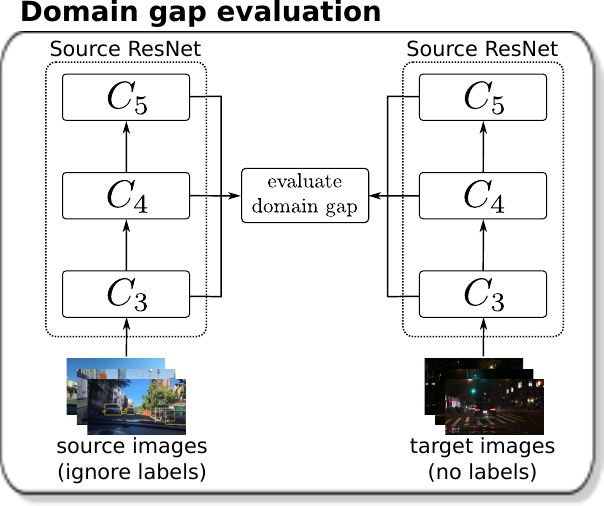}
			\label{fig:domain_gap}
		}
	\caption{An overview of our approach. (a) the RetinaNet is pre-trained on source images with full supervision. (b) The outputs from different blocks $C_3$, $C_4$, and $C_5$ are used to evaluate the domain gap between the source and target images. Note that labels are not needed in this step.}
	\label{fig:approach_overview}
	
\end{figure}
RetinaNet~\citep{retinanet} is adopted as our OD, see Fig.~\ref{fig:pipeline_pretraining}. Specifically, the ResNet backbone~\citep{resnet} has three residual blocks $C_3$, $C_4$, and $C_5$, whose outputs are the inputs of blocks $P_3$, $P_4$, and $P_5$ of the feature pyramid network (FPN)~\citep{fpn}. Then, class and box subnets $B_1$, $B_2$, and $B_3$ receive the outputs of FPN to predict the bounding boxes and object categories.

\section{Domain gap evaluation}
\label{sec:domain_gap_eval}

Let the source images be $\cD_s = \{ x^s_i\}_{i=1}^{N_s}$ and the target images be $\cD_t = \{ x^t_j\}_{j=1}^{N_t}$. 
Firstly, we will train RetinaNet using source images $\cD_s$, see Fig.~\ref{fig:pipeline_pretraining}. Then, to evaluate the domain gap between source domain $\cD_s$ and target domain $\cD_t$, we only use the ResNet backbone, see Fig.~\ref{fig:domain_gap}. 
 
 For an input image $x$, we vectorise the outputs of $C_3$, $C_4$, and $C_5$ and denote them as $C_k(x)$ where $k=3$, $4$, or $5$. In the following sections, we will present three methods for domain gap evaluation: maximum mean discrepancy~\citep{mmd}, sliced Wasserstein distance~\citep{villani2009optimal}, and distance of second-order statistics~\citep{sun2016return}, and show how to apply these methods to the pipeline in Fig.~\ref{fig:domain_gap}.

 \subsection{Maximum mean discrepancy (MMD)}
MMD~\citep{mmd} is a non-parametric metric for estimating the distributional gap between two point sets. It has a variety of applications, e.g., domain adaptation~\citep{yan2019weighted}, deep generative models~\citep{dziugaite2015training}, anomaly detection~\citep{zhang2021unsupervised}, etc. Mathematically, MMD is formulated as follows
\begin{align}
    \text{MMD}(\cD_s, \cD_t) = \begin{Vmatrix} \textstyle\frac{1}{N^s}\sum_{i=1}^{N_s} \phi(x_i^s) - \textstyle\frac{1}{N^t} \sum_{j=1}^{N_t} \phi(x_j^t)\end{Vmatrix}^2,
    \label{eq:mmd}
\end{align}
where, $\phi$ is a feature map. To apply Eq.~\eqref{eq:mmd} to the pipeline in Fig.~\ref{fig:domain_gap}, we consider $\phi(\cdot)$ as $C_k(\cdot)$.
\begin{align}
    \text{MMD}_k(\cD_s, \cD_t) = \begin{Vmatrix}\textstyle\frac{1}{N^s}\sum_{i=1}^{N_s} C_k(x_i^s) - \textstyle\frac{1}{N^t} \sum_{j=1}^{N_t} C_k(x_j^t)\end{Vmatrix}^2,
    \label{eq:our_mmd_1}
\end{align}
where $k = 3$, $4$, or $5$.

To obtain a single value to represent MMD between $\cD_s$ and $\cD_t$, we compute the mean of $\text{MMD}_k$
\begin{align}
    \text{MMD}(\cD_s, \cD_t) =
    \textstyle\frac{1}{3}\sum_{k=3}^5\text{MMD}_k(\cD_s, \cD_t)
    \label{eq:our_mmd_final}
\end{align}

\subsection{Sliced Wasserstein distance (SWD)}
\label{sec:swd}

Wasserstein distance (WD) is a popular metric in optimal transport theory~\citep{villani2009optimal} and have been receiving an increasing attention from the community in designing loss functions for deep generative models~\citep{wgan,adler2018banach,liu2019wasserstein} and domain adaptation~\citep{flamary2016optimal,damodaran2018deepjdot,xu2019wasserstein}. Compared to other measures (e.g.,  Jensen-Shannon
divergence, Kullback-Leibler divergence, and total variation distance), WD takes into account the underlying geometry of probability space. SWD is a variance of WD, which aims to deal with high-dimensional data. Particularly, let empirical distributions of $\cD_s$ and $\cD_t$ as 
\begin{align}
        \mu_s = \textstyle\sum_{i=1}^{N_s}p_i^s \delta_{x_i^s}, \; \; \; \mu_t = \textstyle\sum_{j=1}^{N_t}p^t_j \delta_{x_j^t},
\end{align}
where, $\delta_{x_i^s}$ and $\delta_{x_j^t}$ are the Dirac functions at location $x_i^s$ and $x_j^t$. $p_i^s$ and $p_j^t$ are the probability masses associated to the $i$-th and $j$-th samples. In practice, one typically sets $p_i^s = \textstyle\frac{1}{N_s}$ and $p_j^t = \textstyle\frac{1}{N_t}$. 

Next, define the set 
\begin{align}
    \cB = \begin{Bmatrix} \gamma \in (\bbR^+)^{N_s \times N_t} \; | \; \gamma \bone_{N_t} = \mu_s, \gamma^T \bone_{N_s} = \mu_t\end{Bmatrix},
\end{align}
where, $\bone$ is a vector with all elements equal to 1. 

Then, for each $k=3$, $4$, or $5$, we have the $\text{WD}_k$ as follows 
\begin{align}
    \text{WD}_k(\cD_s, \cD_t) = \min_{\gamma \in \cB}\langle \gamma, \bD_k \rangle_F
\end{align}
where, $\langle \cdot, \cdot \rangle_F$ is the Frobenius dot product and $\bD_k$ is the cost matrix with $\bD_k(i,j) = \begin{Vmatrix} C_k(x_i^s) - C_k(x_j^t)\end{Vmatrix}_2^2$.

However, computing $\bD_k$ is expensive as the dimensionality $d_k$ of $C_k(x)$ is very large. Hence, the main idea of SWD is to project $C_k(x)$ to 1-D first, then WD formulation will be applied to the projected 1-D data. 

Define a set $\{ R_{k,m} \}_{m=1}^M$, where $R_{k,m}$ is the $m$-th one-dimensional linear projection sampled from the uniform on the unit sphere of dimension $d_k-1$. Then, for each $k$, SWD is formalised as 
\begin{align}
    \text{SWD}_k(\cD_s, \cD_t) = \textstyle\frac{1}{M} \sum_{m=1}^M \min_{\gamma \in \cB} \langle \gamma, \hat{\bD}_k \rangle_F
\end{align}
where, $\hat{\bD}_k(i,j) = \begin{Vmatrix} \left(R_{k,m}\right)^T C_k(x_i^s) - \left(R_{k,m}\right)^T C_k(x_j^t)\end{Vmatrix}_2^2$.

Finally, to obtain a scalar to represent the domain gap between $\cD_s$ and $\cD_t$, we compute
\begin{align}
    \text{SWD}(\cD_s, \cD_t) = \textstyle\frac{1}{3} \sum_{k=3}^5 \text{SWD}_k(\cD_s, \cD_t)
\end{align}

\subsection{Distance of second-order statistics (DSS)}
DSS is the distance of covariances of source and target features. It is a popular metric in domain adaptation~\citep{sun2016return,sun2016deep,wang2017deep}. Specifically, suppose we have 
\begin{align*}
    \bC^s_k = [C_k(x^s_1), \dots, C_k(x^s_{N_s})]^T, \; \; \;
    \bC^t_k = [C_k(x^t_1), \dots, C_k(x^t_{N_t})]^T,
\end{align*}
with $k = 3$, $4$, or $5$, i.e., $\bC^s_k(i,j)$ $\begin{pmatrix}\text{or } \bC^t_k(i,j)\end{pmatrix}$ indicates $j$-th feature dimension of source (or target) sample $i$-th. Then, we have the covariance matrices of source and target domains
\begin{align}
    \bK_k^s = \textstyle\frac{1}{N_s-1}\left(
        \left( \bC^s_k \right)^T . \bC^s_k - \frac{1}{N_s}(\bone^T \bC^s_k)^T . (\bone^T \bC^s_k)\right) \\
    \bK_k^t = \textstyle\frac{1}{N_t - 1}\left (
        \left( \bC^t_k \right) ^T . \bC^t_k - \frac{1}{N_t}(\bone^T \bC^t_k)^T . (\bone^T \bC^t_k)\right )
\end{align}
where, $\bone$ is a vector with all elements equal to 1. Finally, we compute the DSS for each $k$
\begin{align}
    \text{DSS}_k = \textstyle\frac{1}{4{d_k}^2}\begin{Vmatrix}
        \bK_k^s - \bK_k^t
    \end{Vmatrix}^2_F,
\end{align}
where, $d_k$ is the dimension of $C_k(x)$.

However, as $d_k$ in the ResNet architecture is very large (i.e., millions of dimensions), computing $\bK_k^s$ and $\bK_k^t$ is intractable. 

To address this issue, we adopt the idea of SWD (see Sec.~\ref{sec:swd}) to project $C_k(x)$ to 1-D first, and then DSS formulation will be applied to the projected 1-D data. Specifically, given a set of one-dimensional linear projections $\{ R_{k,m} \}_{m=1}^M$ randomly sampled from the uniform on the unit sphere of dimension $d_k-1$, for each projection $m$, let
\begin{align}
    \hat{C}_{k,m}(x) = R_{k,m}^T.C_k(x).
\end{align}
Next, we form
\begin{align}
    \hat{\bC}^s_{k,m} = [\hat{C}_{k,m}(x^s_1), \dots, \hat{C}_{k,m}(x^s_{N_s})]^T, \\
    \hat{\bC}^t_{k,m} = [\hat{C}_{k,m}(x^t_1), \dots, \hat{C}_{k,m}(x^t_{N_t})]^T
\end{align}
Covariance matrices of projected features are then computed as
\begin{align}
    \hat{\bK}_{k,m}^s = \textstyle\frac{1}{N_s-1}\left(
        \left(\hat{\bC}^s_{k,m} \right)^T \hat{\bC}^s_{k,m} - \frac{1}{N_s}(\bone^T \hat{\bC}^s_{k,m})^T . (\bone^T \hat{\bC}^s_{k,m}) \right) \\
    \hat{\bK}_{k,m}^t = \textstyle\frac{1}{N_t-1}\left(
        \left(\hat{\bC}^t_{k,m} \right)^T \hat{\bC}^t_{k,m} - \frac{1}{N_t}(\bone^T \hat{\bC}^t_{k,m})^T . (\bone^T \hat{\bC}^t_{k,m}) \right)
\end{align}
 As $\hat{C}_{k,m}(x)$ is 1-D projected feature of $C_k(x)$, computing $\hat{\bK}_{k,m}^s$ and $\hat{\bK}_{k,m}^t$ is fast. 
 
 Finally, the DSS for each $k$ is computed as follows
\begin{align}
    \hat{\text{DSS}}_k = \textstyle\frac{1}{4M}\sum_{m=1}^M\begin{Vmatrix}
        \hat{\bK}_{k,m}^s - \hat{\bK}_{k,m}^t
    \end{Vmatrix}^2_F
\end{align}

As we need a scalar to represent the domain gap between $\cD_s$ and $\cD_t$, we calculate the mean of $\hat{\text{DSS}}_k$
\begin{align}
    \text{DSS}(\cD_s, \cD_t) =
    \textstyle\frac{1}{3}\sum_{k=3}^5\hat{\text{DSS}}_k
    \label{eq:our_dss_final}
\end{align}

\section{Application of domain gap evaluation in continual domain adaptation}
\label{sec:application_domain_gap}
\begin{figure*}[ht]\centering
    \centering    
    \includegraphics[width=0.9\textwidth]{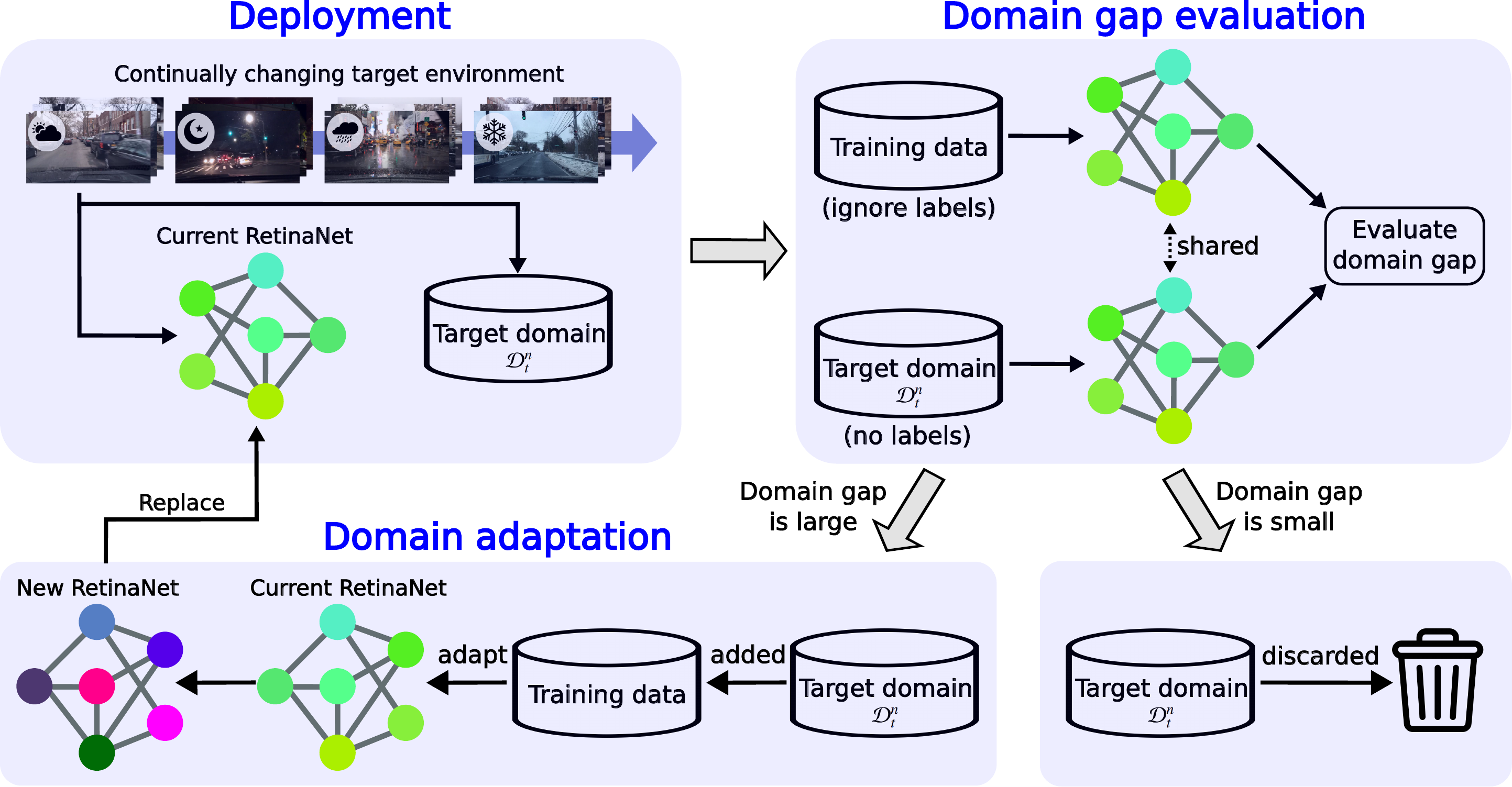}		
    \caption{The application of domain gap evaluation in continually adapting RetinaNet. Specifically, for each target domain $\cD^n_t$, the current RetinaNet is used to evaluate the domain gap between the current training data and $\cD^n_t$ (without the need of labels, as outlined in Sec.~\ref{sec:domain_gap_eval}). If the domain gap is found to be small, $\cD^n_t$ is discarded, and no adaptation is needed. However, if the domain gap is found to be large, $\cD^n_t$ is added to the current training database to refine the current RetinaNet, resulting in a new RetinaNet model which replaces the current RetinaNet in the operating environment.}
    \label{fig:application}
\end{figure*}
Continual domain adaptation is defined as follows: Given a fixed task, RetinaNet is required to adapt to a sequence of $N$ target domains $\left\{\cD^n_t\right\}_{n=1}^N$. Fig.~\ref{fig:application} illustrates our pipeline for applying domain gap evaluation in the continual domain adaptation of RetinaNet. Specifically, given a new target domain $\cD^n_t$, we will determine if domain adaptation is necessary. This is performed by measuring the domain gap between the current training data and the target domain $\cD^n_t$ using the current RetinaNet, as described in Sec.~\ref{sec:domain_gap_eval}. Note that labels are not required for domain gap evaluation.

If the domain gap is found to be smaller than a predefined threshold, the target domain $\cD^n_t$ is discarded as it is deemed unnecessary. This is because if the domain gap between $\cD^n_t$ and the training data is small, the two datasets likely share similar distributions. Therefore, adapting the RetinaNet with this $\cD^n_t$ will not significantly improve its performance and will consume unnecessary resources.

However, if the domain gap is found to be larger than the threshold, $\cD^n_t$ is added to the training database and used to adapt the current RetinaNet. There are a wide range of methods for adapting object detection models in the literature, as outlined in Sec.~\ref{sec:related_work}. Once adapted, the new RetinaNet model replaces the current model in operation.

\section{Experiments}
\subsection{Datasets}
As the primary focus of this paper is on continually adapting ODs to changing environmental conditions, we require object detection datasets that exhibit a variety of conditions. For this reason, we decide to use the DGTA~\citep{DGTA}, BDD~\citep{yu2020bdd100k}, and KITTI~\citep{kitti} datasets in our experiments.

\begin{figure}[ht]\centering
	\centering
	
    \subfloat[][]
	{
		\includegraphics[width=0.4\textwidth]{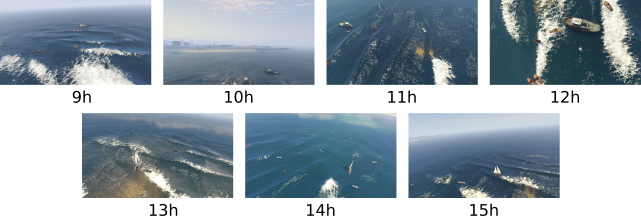}
		\label{fig:dgta_samples:source}
	}
		
	\subfloat[][]
	{
		\includegraphics[width=0.4\textwidth]{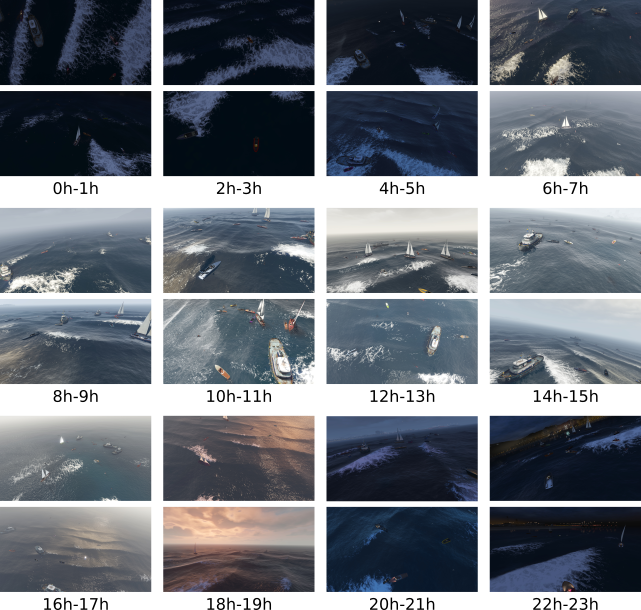}
		\label{fig:dgta_samples:target}
	}
	\caption{Samples of DGTA dataset, where (a) source domain in clear-9h-15h and (b) target domains in overcast in 0h-1h, 2h-3h, 4h-5h, $\dots$, 22h-23h }
	\label{fig:dgta_samples}
\end{figure}

The DGTA dataset~\citep{DGTA} is a synthetic dataset captured from the video game Grand Theft Auto~V~\citep{gtav} and contains five categories, but only the ``boat'' category is used in this experiment. The source data is made up of images captured during clear conditions between 9am and 3pm (denoted as \textbf{clear-9h-15h}). To create the target data, the overcast condition is chosen, and images are divided into 12 chunks, each covering a 2-hour time range (0h-1h, 2h-3h, 4h-5h, $\dots$, 22h-23h). Each chunk forms a different target dataset, resulting in 12 target datasets in total. These 12 target datasets are denoted as \textbf{0h-1h}, \textbf{2h-3h}, \textbf{2h-3h}, $\dots$, \textbf{22h-23h}. Table~\ref{tab:dgta_stats} shows the statistics of the obtained datasets, while Fig.~\ref{fig:dgta_samples} illustrates some samples of the source and target datasets.



\begin{table}[]
    \footnotesize
    \centering
    \begin{tabular}{lcccc}
        \toprule
        \multirow{2}{*}{\textbf{Domain}}& \multicolumn{2}{c}{\textbf{Training}} & \multicolumn{2}{c}{\textbf{Testing}}\\
        \cmidrule(lr){2-3} \cmidrule(lr){4-5}
          &  \textbf{\# imgs} & \textbf{\# anns} & \textbf{\# imgs} & \textbf{\# anns} \\
          \midrule
         clear 9h-15h & 4,658  & 48,643 &  1,165 & 12,003 \\ 
         overcast 0h-1h & 1,826  & 17,247 & 457 & 3,904 \\
         overcast 2h-3h &  2,038 & 18,190 & 510 & 4,428 \\
         overcast 4h-5h &  1,595 & 16,212 & 399 & 4,115 \\
         overcast 6h-7h &  1,549 & 17,279 & 388 & 4,324 \\
         overcast 8h-9h &  1,545 & 16,037 & 387 & 3,801\\
         overcast 10h-11h &  1,678 & 18,828 & 420 & 4,817 \\
         overcast 12h-13h & 1,436 & 14,776 & 359 & 3,571 \\
         overcast 14h-15h & 1,308 & 12,430 & 327 & 3,139\\
         overcast 16h-17h & 1,368 & 12,987 & 343 & 3392 \\
         overcast 18h-19h & 1,708 & 19,689 & 428 & 4,871 \\
         overcast 20h-21h & 1,639 & 17,490 & 410 & 4,275 \\
         overcast 22h-23h & 1,784 & 17,884 & 447 & 4,215 \\
         \bottomrule
         
    \end{tabular}
    \caption{Statistics of DGTA, where ``\#~imgs'' and ``\#~anns'' are abbreviations of ``number of images'' and ``number of annotations''.}
    \label{tab:dgta_stats}
\end{table}

BDD~\citep{yu2020bdd100k} is a real dataset containing 100,000 driving videos under different weather conditions (clear, foggy, overcast, cloudy, rainy, and snowy) and time of day (daytime and night). The source domain is selected from the \textbf{clear-daytime} condition, while the target domains are selected from \textbf{clear-night}, \textbf{cloudy-daytime}, \textbf{overcast-daytime}, \textbf{rainy-daytime}, \textbf{rainy-night}, \textbf{snowy-daytime}, and \textbf{snowy-night} conditions. Remaining conditions are not considered due to the small number of images. Then, two following settings are created for BDD dataset.
\vspace{-\topsep}
\begin{itemize}
    \setlength{\parskip}{0pt}
	\setlength{\itemsep}{0pt plus 1pt}
    \item Single-class BDD (denoted as \textbf{sBDD}): only the ``car'' category is used.
    \item Multi-class BDD (denoted as \textbf{mBDD}): four categories (pedestrian, car, traffic light, traffic sign) are used.
\end{itemize}
Tables~\ref{tab:sbdd_stats} and~\ref{tab:mbdd_stats} respectively present the statistics of sBDD and mBDD, while Fig.~\ref{fig:bdd_samples} illustrates their samples.

\begin{figure}[ht]\centering
	\centering
    \includegraphics[width=0.4\textwidth]{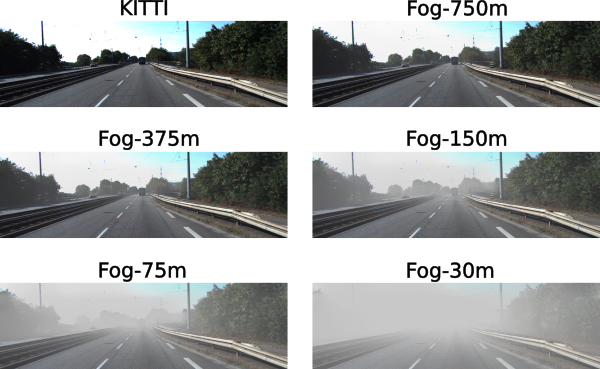}
    \label{fig:kitti_samples}
	\caption{Samples of KITTI benchmark, where source domain is KITTI images, and target domains are Fog-750m/375m/150m/75m/30m.} 
	\label{fig:kitti_samples}
\end{figure}

KITTI~\citep{kitti} is a widely used dataset in autonomous driving, which contains seven object categories, but only the ``car" category is used in this experiment. The source domain is composed of KITTI images, while the target domains are generated using the technique of~\citep{kittifog}, resulting in five different levels of fog: \textbf{fog-750m}, \textbf{fog-375m}, \textbf{fog-150m}, \textbf{fog-75m}, and \textbf{fod-30m}; see Fig.~\ref{fig:kitti_samples} for sample images. The training and testing sets are randomly split, leading to 3,740 training images with 14,655 annotations and 3,741 testing images with 14,087 annotations.

\begin{table}[]
    \footnotesize
    \centering
    \begin{tabular}{lcccc}
        \toprule
         \multirow{2}{*}{\textbf{Domain}} & \multicolumn{2}{c}{\textbf{Training}} & \multicolumn{2}{c}{\textbf{Testing}}  \\
         \cmidrule(lr){2-3} \cmidrule(lr){4-5}
         & \textbf{\# imgs} & \textbf{\# anns} & \textbf{\# imgs} & \textbf{\# anns} \\
         \midrule
         clear-daytime & 12,347 & 138,754 & 1,750 & 19,745 \\
         clear-night & 22,428 & 198,137 & 3,222 & 30,350 \\
         cloudy-daytime & 4,242 & 48,649 & 632 & 7,432 \\
         overcast-daytime & 7,501 & 89,463 & 1,032 & 12,453 \\
         rainy-daytime & 2,497 & 24,851 & 393 & 3,958 \\
         rainy-night & 2,181 & 18,002 & 281 & 2,372 \\
         snowy-daytime & 2,835 & 28,040 & 418 & 4,477 \\
         snowy-night & 2,214 & 19,092 & 266 & 2,247 \\
         \bottomrule
         
    \end{tabular}
    \caption{Statistics of sBDD, where ``\#~imgs'' and ``\#~anns'' are abbreviations of ``number of images'' and ``number of annotations''.}
    \label{tab:sbdd_stats}
\end{table}

\begin{table}[]
    \centering
    \footnotesize
    \begin{tabular}{lcccc}
        \toprule
         \multirow{2}{*}{\textbf{Domain}} & \multicolumn{2}{c}{\textbf{Training}} & \multicolumn{2}{c}{\textbf{Testing}}  \\
         \cmidrule(lr){2-3} \cmidrule(lr){4-5}
         & \textbf{\# imgs} & \textbf{\# anns} & \textbf{\# imgs} & \textbf{\# anns} \\
         \midrule
         clear-daytime & 12,450 & 225,523 & 1,764 & 32,331 \\
         clear-night &  22,871 & 354,717 & 3,274 & 52,910 \\
         cloudy-daytime & 4,261 & 80,261 & 638 & 12,264  \\
         overcast-daytime & 7,550 & 152,106 & 1,039 & 20,840 \\
         rainy-daytime & 2,521 & 44,954 & 396 & 7,007  \\
         rainy-night &  2,208 & 33,384 & 286 & 4,564 \\
         snowy-daytime & 2,861 & 50,992 & 422 & 7,726  \\
         snowy-night & 2,248 & 35,588 & 273 & 4,383 \\
         \bottomrule
         
    \end{tabular}
    \caption{Statistics of mBDD, where ``\#~imgs'' and ``\#~anns'' are abbreviations of ``number of images'' and ``number of annotations''.}
    \label{tab:mbdd_stats}
\end{table}

\begin{figure}[ht]\centering
	\centering
    \includegraphics[width=0.4\textwidth]{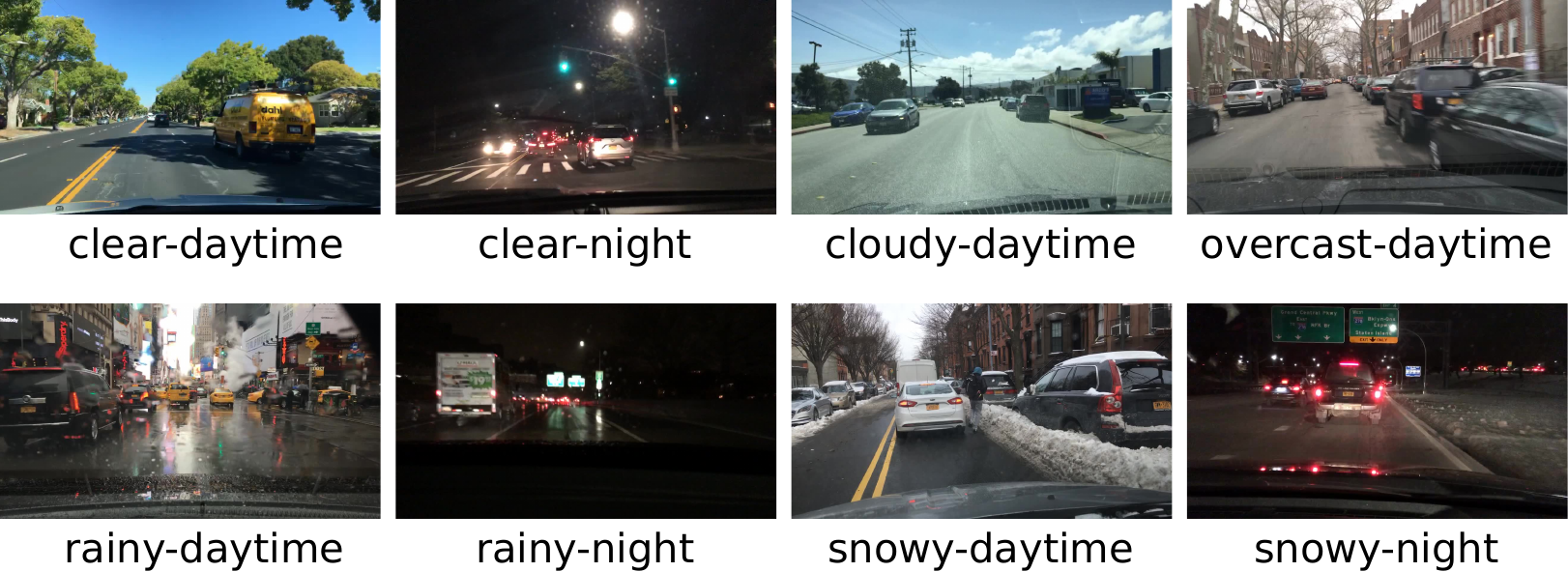}
    \label{fig:bdd_samples}
	\caption{Samples of BDD dataset, where source domain is clear-daytime, and target domains are clear-night, cloudy-daytime, overcast-daytime, rainy-daytime, rainy-night, snowy-daytime, and snowy-night. }
	\label{fig:bdd_samples}
\end{figure}

\begin{figure*}[h]
	\centering
	\includegraphics[width=0.95\textwidth]{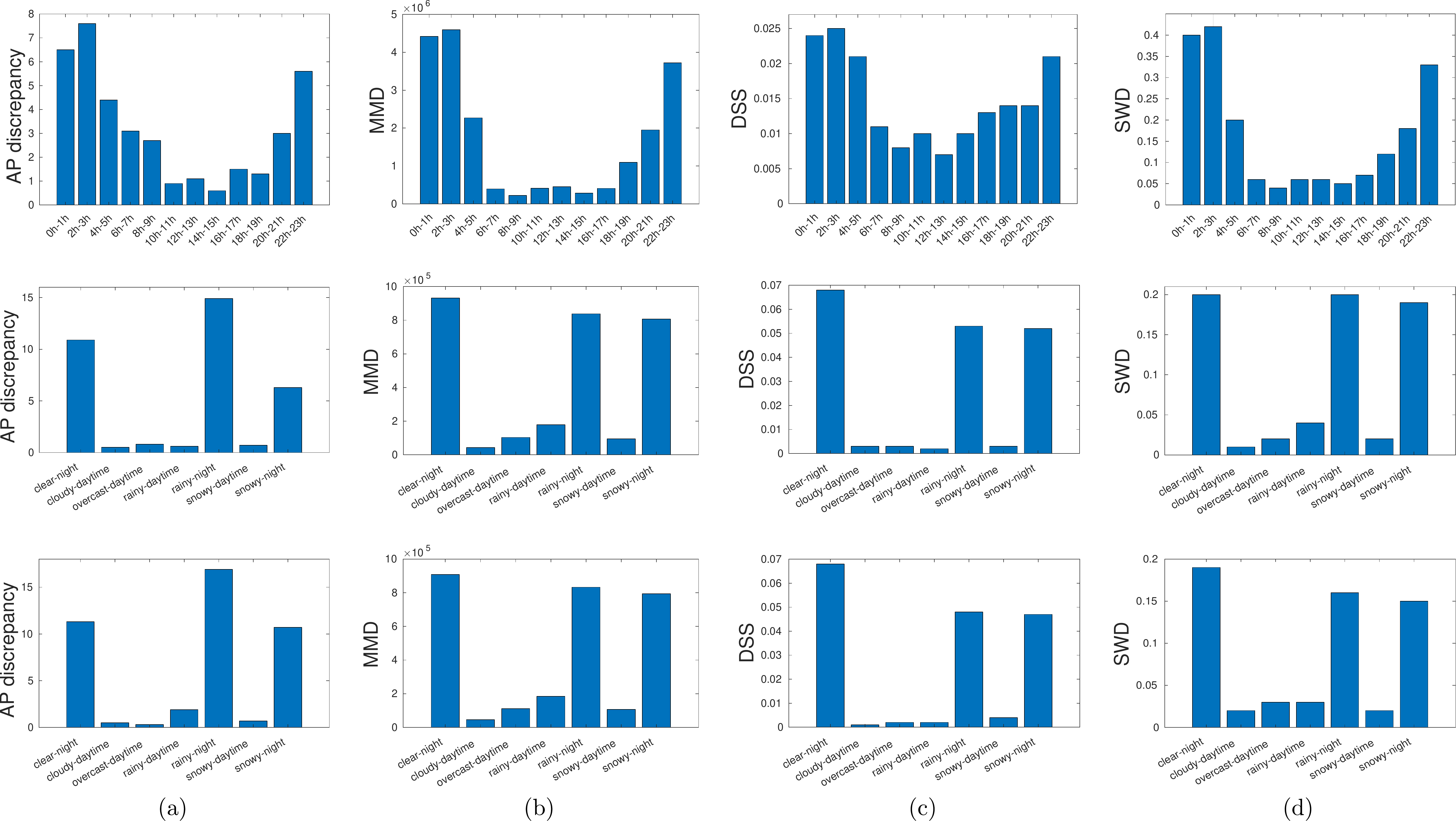}
    \caption{Histograms of AP discrepancy and domain gap in DGTA (1st row), sBDD (2nd row) and mBDD (3rd row) datasets, where (a) AP discrepancy, (b)-(d) domain gap between source and target domains evaluated by MMD, DSS and SWD. The source model is trained on source data \textbf{clear-9h-15h} (for DGTA) and \textbf{clear-daytime} (for sBDD and mBDD), and target domains are specified in x-axis values.}
    \label{fig:correlation}
\end{figure*}

\subsection{Experimental setup} \label{sec:exp_setup}

RetinaNet with a ResNet-50 backbone is trained on the source domain (i.e., \textbf{clear-9h-15h} for DGTA and \textbf{clear-daytime} for sBDD and mBDD) to obtain a source OD. Specifically, the OD is initialised from the OD pretrained on COCO~\citep{coco}. We train the OD with a learning rate of $0.0002$ for $30$k iterations, then reduce the learning rate to $0.00002$ for another $30$k iterations. Unless stated otherwise, the batch size of $4$, momentum of $0.9$, weight decay of $0.001$, and $M=10$ are used in the experiment. 

The performance of the detection is evaluated based on the average precision and average recall metrics, both measured with a threshold of 50\%. These metrics are abbreviated as AP50 and AR50, respectively. Each experiment is conducted in an NVIDIA~GeForce~RTX~2080~Ti and implemented using Detectron2~\citep{wu2019detectron2}.

\subsection{Correlation between domain gap and detection accuracy}
\begin{table}
    \centering
    \begin{tabular}{ lccc }
     \toprule
      & \textbf{DGTA} & \textbf{sBDD} & \textbf{mBDD} \\
      \midrule
      MMD & 0.15 & 0.08 & 0.05 \\
      DSS & 0.08 & 0.06 & 0.08 \\
      SWD & 0.09 & 0.07 & 0.08 \\
     \bottomrule
    \end{tabular}
    \caption{Kullback–Leibler divergence between domain gap and AP discrepancy}
    \label{tab:KL}
\end{table}

\subsubsection{Setup} Inspired by~\citep[Theorem~2]{ben2010theory} stating that the discrepancy of source and target errors are upper-bounded by the domain gap, we firstly define the discrepancy of AP 
\begin{align}
\text{AP discrepancy} = \left |\text{AP}_{\text{source}} - \text{AP}_{\text{target}} \right|
\end{align}
where, $\text{AP}_{\text{source}}$ and $\text{AP}_{\text{target}}$ are APs of source OD on source and target testing sets. Then, the domain gap will be evaluated by the three metrics presented in Sec.~\ref{sec:domain_gap_eval}.

\subsubsection{Results} The correlation between the AP discrepancy and domain gap is shown in Fig.~\ref{fig:correlation}. Specifically, when the gap between the source and target domains is large, the AP discrepancy will also be large, meaning that the OD trained on the source domain will perform poorly in the target domain. Conversely, when the gap is small, the source OD will deliver an accuracy on the target domain that is similar to its accuracy on the source domain

To quantitatively compare the performance of MMD, DSS, and SWD, we compute the Kullback–Leibler divergence between the histograms of domain gap and AP discrepancy, as presented in Table~\ref{tab:KL}. The results show that DSS and SWD perform consistently well across the DGTA, sBDD, and mBDD datasets, while MMD performs slightly worse on the DGTA dataset. This is likely because MMD only considers first-order statistics, while DSS considers second-order statistics and SWD takes into account data geometry.

\begin{figure}
    \centering
    \includegraphics[width=0.4\textwidth]{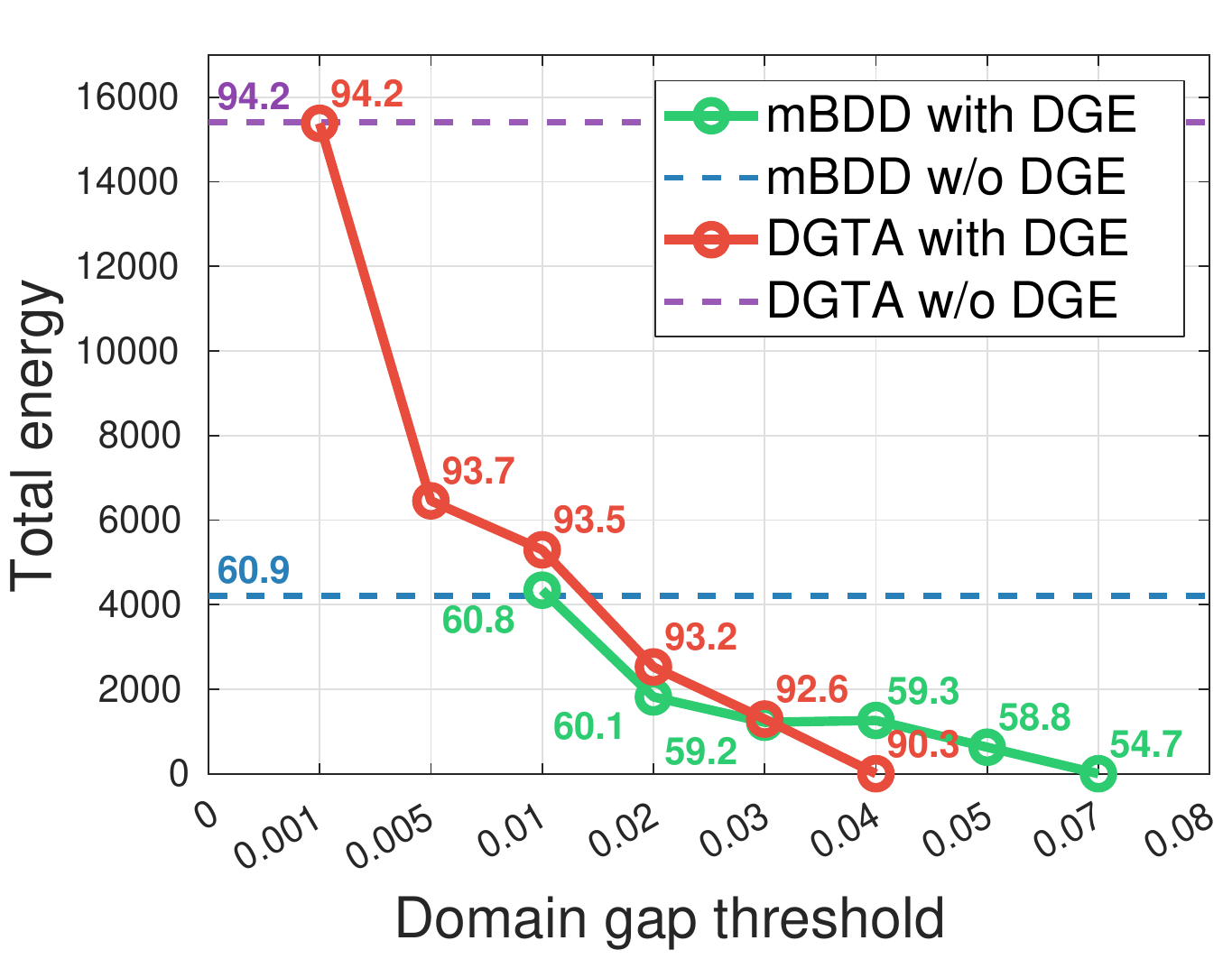}
    \caption{Correlation between total energy consumption for continual domain adaptation, the domain gap threshold, and AP50, where AP50 is reported adjacent to the markers~\tikzcircle[myred]{3pt} and~\tikzcircle[mygreen]{3pt}}.
    \label{fig:diff_thr}
\end{figure}

\begin{table*}
    \scriptsize
    \centering
    \begin{tabular}{ l P{3.5em} P{3.5em} P{3.5em} P{3.5em} P{3.5em} P{3.5em} P{3.5em} P{3.5em} P{3.5em} P{3.5em} P{3.5em} P{3.5em}  P{3em}  P{3em}}
     \toprule
      \multirow{3}{*}{\textbf{Time}} & \multicolumn{12}{l}{ \begin{tikzpicture} \node (t) at (0, 0) {$\mathbf{t}$}; \draw [-To,line width=0.3mm](0.1,0) -- (14.9,0);\end{tikzpicture}} & \multirow{3}{*}{\textbf{Mean}} \\
      
       &  \textbf{0h-1h} & \textbf{2h-3h} & \textbf{4h-5h} & \textbf{6h-7h} & \textbf{8h-9h} & \textbf{10h-11h} & \textbf{12h-13h} & \textbf{14h-15h} & \textbf{16h-17h} & \textbf{18h-19h} & \textbf{20h-21h} & \textbf{22h-23h} & \\
      \midrule
      \multirow{2}{*}{No DA} & 86.1$\pm$0.3 & 85.0$\pm$0.1 & 88.6$\pm$0.1 & 90.2$\pm$0.2 & 90.8$\pm$0.1 & 94.1$\pm$0.1 & 94.5$\pm$0.1 & 92.9$\pm$0.2 & 91.7$\pm$0.2 & 92.1$\pm$0.1 & 90.2$\pm$0.1 & 87.1$\pm$0.2 & 90.3$\pm$0.1 \\
      & 91.6$\pm$0.4 & 90.5$\pm$0.2 & 92.5$\pm$0.1 & 93.1$\pm$0.2 & 93.6$\pm$0.1 & 96.3$\pm$0.1 & 96.7$\pm$0.1 & 95.0$\pm$0.1 & 94.0$\pm$0.2 & 94.7$\pm$0.1 & 93.6$\pm$0.2 & 92.0$\pm$0.2 & 93.6$\pm$0.1 \\
      \midrule
      \multirow{2}{*}{SDA w/o DGE}  & 94.2$\pm$0.1  &  94.5$\pm$0.1 & 93.5$\pm$0.2 & 93.1$\pm$0.2 & 93.3$\pm$0.1 & 95.2$\pm$0.2 & 95.5$\pm$0.2 & 94.7$\pm$0.3 & 94.2$\pm$0.2 & 94.2$\pm$0.3 & 93.5$\pm$0.4 & 94.9$\pm$0.1 & 94.2$\pm$0.1 \\
      & 96.4$\pm$0.2 &  96.5$\pm$0.1 & 95.9$\pm$0.2 & 95.5$\pm$0.2 & 95.5$\pm$0.1 & 97.0$\pm$0.1 & 97.0$\pm$0.1 & 96.5$\pm$0.3 & 96.1$\pm$0.3 & 96.3$\pm$0.2 & 95.6$\pm$0.3 & 96.6$\pm$0.1 & 96.3$\pm$0.1 \\
      \midrule
      \multirow{2}{*}{SDA with DGE} & 94.3$\pm$0.1  & 94.6$\pm$0.1 & 92.6$\pm$0.3 & 90.8$\pm$0.2 & 91.3$\pm$0.2 & 94.5$\pm$0.1 & 94.9$\pm$0.1 & 93.3$\pm$0.1 & 92.1$\pm$0.3 & 92.4$\pm$0.2 & 92.7$\pm$0.3 & 94.2$\pm$0.3 & 93.1$\pm$0.2 \\
      
      & 96.5$\pm$0.2 & 96.7$\pm$0.1 & 95.1$\pm$0.3 & 93.7$\pm$0.2 & 94.1$\pm$0.1 & 96.5$\pm$0.2 & 96.7$\pm$0.1 & 95.3$\pm$0.2 & 94.3$\pm$0.3 & 95.0$\pm$0.2 & 95.1$\pm$0.2 & 96.2$\pm$0.2 & 95.4$\pm$0.2\\
      \midrule
      \multirow{2}{*}{UDA w/o DGE}  & 89.1$\pm$0.8 & 89.5$\pm$0.5  & 89.8$\pm$0.7  & 90.9$\pm$0.2 & 91.3$\pm$0.8 & 94.3$\pm$0.2 & 94.9$\pm$0.1 & 93.5$\pm$0.1  & 92.5$\pm$0.5  & 92.1$\pm$0.5  & 90.5$\pm$0.4  & 89.0$\pm$0.4 & 91.5$\pm$0.1 \\
      & 92.9$\pm$0.8 & 93.1$\pm$0.5 & 92.9$\pm$0.6 & 93.9$\pm$0.3 & 94.1$\pm$0.8 & 96.3$\pm$0.2 & 96.8$\pm$0.1 & 95.7$\pm$0.2 & 94.6$\pm$0.4 & 94.8$\pm$0.5 & 93.6$\pm$0.5 & 92.5$\pm$0.3 & 94.3$\pm$0.1 \\
      \midrule
       \multirow{2}{*}{UDA with DGE} &  89.4$\pm$0.8 & 89.2$\pm$0.8 & 89.8$\pm$0.4  & 91.0$\pm$0.4  & 91.2$\pm$0.3  & 94.1$\pm$0.3  & 94.6$\pm$0.1  & 93.2$\pm$0.3  & 92.3$\pm$0.2  & 92.1$\pm$0.2  & 90.9$\pm$0.3  & 90.0$\pm$0.5 & 91.5$\pm$0.2 \\
       & 93.2$\pm$0.4 & 93.0$\pm$0.5 & 93.0$\pm$0.3 & 94.1$\pm$0.2 & 94.2$\pm$0.3 & 96.3$\pm$0.3 & 96.6$\pm$0.1 & 95.3$\pm$0.2 & 94.5$\pm$0.2 & 94.9$\pm$0.1 & 94.0$\pm$0.3 & 93.3$\pm$0.4 & 94.4$\pm$0.2 \\
      \bottomrule
    \end{tabular}
    \caption{Comparison between AP50 (top) and AR50 (bottom) in continual domain adaptation on DGTA, where target domain is sequentially changed \textbf{0h-1h} $\rightarrow$ \textbf{1h-2h} $\rightarrow$ \dots $\rightarrow$ \textbf{22h-23h}.}
    \label{tab:DGTA_results}
\end{table*}

In the following section, we will use DSS to evaluate the benefit of domain gap evaluation in continual domain adaptation of object detectors~\footnote{MMD and SWD are expected to deliver a similar outcome as DSS}.

\subsection{Continual domain adaptation of RetinaNet}

\subsubsection{Setup} To simulate a continual change of environmental conditions, we consider the following configuration
\vspace{-\topsep}
\begin{itemize}
    \setlength{\parskip}{0pt}
	\setlength{\itemsep}{0pt plus 1pt}
    \item For DGTA, the source domain is still \textbf{clear-9h-15h}, but the target domain in overcast will be sequentially changed \textbf{0h-1h} $\rightarrow$ \textbf{1h-2h} $\rightarrow$ \textbf{2h-3h} $\rightarrow$ \dots $\rightarrow$ \textbf{22h-23h}.
    \item For sBDD and mBDD, the source domain is \textbf{clear-daytime}, but the target domain will be continually changed \textbf{clear-night} $\rightarrow$ \textbf{cloudy-daytime} $\rightarrow$ \textbf{overcast-daytime} $\rightarrow$ \textbf{rainy-daytime} $\rightarrow$ \textbf{rainy-night} $\rightarrow$ \textbf{snowy-daytime} $\rightarrow$ \textbf{snowy-night}.
\end{itemize}
If the OD encounters a new condition, we will adapt the OD with the following domain adaptation methods
\vspace{-\topsep}
\begin{itemize}
    \setlength{\parskip}{0pt}
	\setlength{\itemsep}{0pt plus 1pt}
    \item \textbf{No domain adaptation} (\texttt{NoDA}): The model is only trained on source domain and will never be adapted.
    \item \textbf{Supervised domain adaptation} (\texttt{SDA}): Annotations are assumed to be fully available, thus the OD can be fine-tuned in a supervised manner. This provides an upper-bound performance for domain adaptation.
    \item \textbf{Unsupervised domain adaptation} (\texttt{UDA}): Labels are only available in the source data, and the new data is added to the unlabelled target set. Specifically, we use adversarial RetinaNet~\citep{PASQUALINO2021104098} for the adaptation.
\end{itemize}
In all adaptation methods, the current model is used as the initialisation, then the model is trained for $20$k iterations with a fixed learning rate $0.0002$. Also, we consider following settings
\begin{table*}
    \centering
        \begin{tabular}{ l c c c c c c }
             \hline
              \multirow{3}{*}{\textbf{Time}} & \multicolumn{5}{l}{ \begin{tikzpicture} \node (t) at (0, 0) {$\mathbf{t}$}; \draw [-To,line width=0.3mm](0.1,0) -- (8.5,0);\end{tikzpicture}} & \multirow{3}{*}{\textbf{Mean}} \\
              
               &  \textbf{fog-750m} & \textbf{fog-375m} & \textbf{fog-150m} & \textbf{fog-75m} & \textbf{fog-30m}   \\
              \midrule
              \multirow{2}{*}{No DA} & 89.1$\pm$0.1 & 86.6$\pm$0.1 & 79.3$\pm$0.1 & 66.6$\pm$0.5 & 35.0$\pm$0.6 & 71.3$\pm$0.2\\
              & 94.7$\pm$0.1 & 92.5$\pm$0.1 & 85.5$\pm$0.2 & 72.6$\pm$0.6 & 37.3$\pm$0.6 & 76.5$\pm$0.2\\
              \midrule
              \multirow{2}{*}{SDA w/o DGE}  & 93.0$\pm$0.1 & 92.9$\pm$0.1 & 92.5$\pm$0.1 & 91.6$\pm$0.1 & 85.6$\pm$0.2 & 91.1$\pm$0.1\\
              & 97.3$\pm$0.1 & 96.9$\pm$0.1 & 96.7$\pm$0.1 & 96.3$\pm$0.3 & 95.5$\pm$0.2 & 96.5$\pm$0.2\\
              \midrule
              \multirow{2}{*}{SDA with DGE} & 89.1$\pm$0.1 & 86.6$\pm$0.1 & 92.0$\pm$0.1 & 91.4$\pm$0.1 & 86.3$\pm$0.1 & 89.1$\pm$0.1 \\
              & 94.7$\pm$0.1 & 92.5$\pm$0.1 & 97.2$\pm$0.1 & 96.6$\pm$0.2 & 96.1$\pm$0.4 & 95.4$\pm$0.2 \\
              \midrule
              \multirow{2}{*}{UDA w/o DGE} & 89.0$\pm$0.4 & 85.8$\pm$0.6 & 79.5$\pm$0.8 & 71.3$\pm$1.0 & 43.6$\pm$1.0 & 73.9$\pm$0.8 \\
              & 93.8$\pm$0.4 & 90.0$\pm$0.6 & 82.9$\pm$1.1 & 74.2$\pm$1.2 & 45.6$\pm$1.1 & 77.3$\pm$0.9 \\
              \midrule
              \multirow{2}{*}{UDA with DGE} & 89.1$\pm$0.1 & 86.6$\pm$0.1 & 80.4$\pm$1.1 & 69.8$\pm$0.9 & 42.1$\pm$0.8 & 73.6$\pm$0.6 \\
              & 94.7$\pm$0.1 & 92.5$\pm$0.1 & 84.9$\pm$1.5 & 73.1$\pm$0.7 & 43.7$\pm$0.9 & 77.8$\pm$0.7\\
              \bottomrule
        \end{tabular}
    \caption{Comparison between AP50 (top) and AR50 (bottom) in continual domain adaptation on KITTI, where target domain is sequentially changed \textbf{fog-750m} $\rightarrow$ \textbf{fog-375m} $\rightarrow$ \textbf{fog-150m} $\rightarrow$ \textbf{fog-75m} $\rightarrow$ \textbf{fog-30m}.}
    \label{tab:kitti}
\end{table*}

\begin{table*}
    \small
    \centering
        \begin{tabular}{ l P{4em} P{4em} P{4em} P{4em} P{4em} P{4em} P{4em}  P{4em} }
             \toprule
              \multirow{3}{*}{\textbf{Time}} & \multicolumn{7}{l}{ \begin{tikzpicture} \node (t) at (0, 0) {$\mathbf{t}$}; \draw [-to, line width=0.3mm](0.1,0) -- (11,0);\end{tikzpicture}} & \multirow{3}{*}{\textbf{Mean}} \\
               &  \textbf{clear-night} & \textbf{cloudy-daytime} & \textbf{overcast-daytime} & \textbf{rainy-daytime} & \textbf{rainy-nigh}t & \textbf{snowy-daytime} & \textbf{snowy-night} &  \\
              \midrule
              \multirow{2}{*}{No DA} & 66.1$\pm$0.1 & 77.3$\pm$0.1 & 77.6$\pm$0.1 & 77.4$\pm$0.1 & 62.0$\pm$0.1 & 77.5$\pm$0.2 & 70.5$\pm$0.2 & 72.6$\pm$0.1 \\ 
              
              & 88.0$\pm$0.1 & 89.8$\pm$0.1 & 90.2$\pm$0.1 & 91.2$\pm$0.1 & 85.2$\pm$0.2 & 90.5$\pm$0.2 & 90.3$\pm$0.1 & 89.3$\pm$0.1 \\
              \midrule
              \multirow{2}{*}{SDA w/o DGE}  & 74.2$\pm$0.1 & 77.5$\pm$0.2 & 78.3$\pm$0.1 & 79.3$\pm$0.2 & 74.3$\pm$0.3 & 78.9$\pm$0.2 & 78.1$\pm$0.5 & 77.2$\pm$0.1 \\
              & 91.5$\pm$0.1 & 90.0$\pm$0.1 & 90.3$\pm$0.2 & 92.2$\pm$0.2 & 91.1$\pm$0.5 & 91.4$\pm$0.1 & 93.4$\pm$0.4 & 91.4$\pm$0.1\\
              \midrule
               \multirow{2}{*}{SDA with DGE} & 74.2$\pm$0.3 & 77.6$\pm$0.1 & 77.7$\pm$0.2 & 77.9$\pm$0.5 & 71.7$\pm$0.5 & 78.3$\pm$0.2 & 77.1$\pm$0.1 & 76.4$\pm$0.2 \\

               & 91.6$\pm$0.3 & 90.0$\pm$0.2 & 90.0$\pm$0.2 & 91.3$\pm$0.3 & 89.9$\pm$0.7 & 91.0$\pm$0.3 & 93.2$\pm$0.4 & 91.0$\pm$0.1\\
              \midrule
              \multirow{2}{*}{UDA w/o DGE}  &  66.5$\pm$0.3 & 77.8$\pm$0.3 & 77.8$\pm$0.3 & 78.8$\pm$0.4 & 62.3$\pm$0.4 & 78.4$\pm$0.3 & 70.8$\pm$0.4 & 73.2$\pm$0.1 \\
              & 88.4$\pm$0.4 & 90.2$\pm$0.2 & 90.3$\pm$0.3 & 91.5$\pm$0.3 & 85.8$\pm$0.4 & 90.8$\pm$0.3 & 90.8$\pm$0.3 & 89.7$\pm$0.1\\
              \midrule
              \multirow{2}{*}{UDA with DGE} & 66.5$\pm$0.2 & 77.7$\pm$0.1 & 77.9$\pm$0.1 & 77.7$\pm$0.2 & 62.0$\pm$0.3 & 77.5$\pm$0.3 & 70.6$\pm$0.3 & 72.8$\pm$0.1 \\
              & 88.3$\pm$0.3 & 90.2$\pm$0.1 & 90.3$\pm$0.1 & 91.4$\pm$0.1 & 85.4$\pm$0.3 & 90.8$\pm$0.3 & 90.6$\pm$0.3 & 89.6$\pm$0.1\\
              \bottomrule
        \end{tabular}
        \caption{Comparison between AP50 (top) and AR50 (bottom) in continual domain adaptation on sBDD, where target domain is sequentially changed \textbf{clear-night} $\rightarrow$ \textbf{cloudy-daytime} $\rightarrow$ \textbf{overcast-daytime} $\rightarrow$ \textbf{rainy-daytime} $\rightarrow$ \textbf{rainy-night} $\rightarrow$ \textbf{snowy-daytime} $\rightarrow$ \textbf{snowy-night}.}
    \label{tab:sbdd}
\end{table*}

\vspace{-\topsep}
\begin{itemize}
    \setlength{\parskip}{0pt}
	\setlength{\itemsep}{0pt plus 1pt}
    \item \textbf{Without domain gap evaluation} (\texttt{w/o DGE}): The model is always adapted in every condition.
    \item \textbf{With domain gap evaluation} (\texttt{with DGE}): Using DSS, we will verify if the domain adaptation is necessary for the new data. 
\end{itemize}

To quantitatively evaluate the adaptation cost, we measure the total GPU energy consumption for domain adaptation. The GPU energy is measured using the pyJoules~\citep{pyjoules}.

\subsubsection{Results} \label{sec:exp_cont_da}

\vspace{0.5em}
\noindent\textbf{Effects of different domain gap thresholds} \,\,\, The experiment is conducted on DGTA and mBDD. \texttt{SDA} is used for domain adaptation. AP50 is reported as the detection accuracy.

\begin{table*}
    \small
    \centering
        \begin{tabular}{ l P{4em} P{4em} P{4em} P{4em} P{4em} P{4em} P{4em} P{4em} }
             \hline
              \multirow{3}{*}{\textbf{Time}} & \multicolumn{7}{l}{ \begin{tikzpicture} \node (t) at (0, 0) {$\mathbf{t}$}; \draw [-To,line width=0.3mm](0.1,0) -- (11,0);\end{tikzpicture}} & \multirow{3}{*}{\textbf{Mean}} \\
              
               &  \textbf{clear-night} & \textbf{cloudy-daytime} & \textbf{overcast-daytime} & \textbf{rainy-daytime} & \textbf{rainy-night} & \textbf{snowy-daytime} & \textbf{snowy-night} &   \\
              \midrule
              \multirow{2}{*}{No DA} & 49.3$\pm$0.1 & 60.0$\pm$0.2 & 60.1$\pm$0.1 & 58.7$\pm$0.1 & 43.7$\pm$0.2 & 60.9$\pm$0.1 & 49.7$\pm$0.3 & 54.6$\pm$0.1 \\

              & 73.4$\pm$0.1 & 76.8$\pm$0.1 & 77.3$\pm$0.1 & 78.4$\pm$0.1 & 69.1$\pm$0.1 & 78.2$\pm$0.1 & 73.9$\pm$0.3 & 75.3$\pm$0.1\\
              \midrule
              \multirow{2}{*}{SDA w/o DGE}  &  58.3$\pm$0.2 & 60.6$\pm$0.4 & 61.8$\pm$0.2 & 61.7$\pm$0.2 & 57.7$\pm$0.2 & 63.6$\pm$0.2 & 62.5$\pm$0.2 & 60.9$\pm$0.1 \\
              & 79.1$\pm$0.2 & 76.8$\pm$0.5 & 78.4$\pm$0.4 & 80.1$\pm$0.5 & 79.6$\pm$0.5 & 80.1$\pm$0.3 & 82.3$\pm$0.2 & 79.5$\pm$0.1 \\
              \midrule
              \multirow{2}{*}{SDA with DGE} & 58.3$\pm$0.4 & 60.4$\pm$0.5 & 61.1$\pm$0.1 & 60.5$\pm$0.3 & 55.4$\pm$0.1 & 63.0$\pm$0.1 & 61.0$\pm$0.5 & 60.0$\pm$0.1 \\
              & 79.3$\pm$0.3 & 76.9$\pm$0.4 & 77.7$\pm$0.2 & 79.7$\pm$0.4 & 78.4$\pm$0.2 & 79.8$\pm$0.3 & 81.4$\pm$0.5 & 79.0$\pm$0.2  \\
              \midrule
              \multirow{2}{*}{UDA w/o DGE} &  49.9$\pm$0.3 & 61.5$\pm$0.1 & 61.4$\pm$0.1 & 59.6$\pm$0.4 & 43.9$\pm0.3$ & 61.1$\pm$0.2 & 49.7$\pm$0.4 & 55.3$\pm$0.1 \\
              
              & 73.9$\pm$0.3 & 78.1$\pm$0.1 & 78.5$\pm$0.1 & 79.1$\pm$0.4 & 69.2$\pm$0.3 & 79.6$\pm$0.2 & 73.5$\pm$0.5 & 75.9$\pm$0.1\\
              \midrule
              \multirow{2}{*}{UDA with DGE} & 49.8$\pm$0.6 & 60.7$\pm$0.3 & 60.8$\pm$0.2 & 59.3$\pm$0.4 & 44.4$\pm$0.5 & 62.0$\pm$0.2 & 50.9$\pm$0.7 & 55.4$\pm$0.4 \\
              & 73.8$\pm$0.6 & 77.6$\pm$0.2 & 78.1$\pm$0.2 & 79.1$\pm$0.4 & 69.9$\pm$0.8 & 79.1$\pm$0.3 & 74.8$\pm$0.3 & 76.1$\pm$0.3 \\
              \bottomrule
        \end{tabular}
    \caption{Comparison between AP50 (top) and AR50 (bottom) in continual domain adaptation on mBDD, where target domain is sequentially changed \textbf{clear-night} $\rightarrow$ \textbf{cloudy-daytime} $\rightarrow$ \textbf{overcast-daytime} $\rightarrow$ \textbf{rainy-daytime} $\rightarrow$ \textbf{rainy-night} $\rightarrow$ \textbf{snowy-daytime} $\rightarrow$ \textbf{snowy-night}.}
    \label{tab:mbdd}
\end{table*}

The result is shown in Fig.~\ref{fig:diff_thr}. When the threshold is set at a relatively low value (i.e., 0.001 for DGTA and 0.01 for mBDD), domain adaptation is carried out under all conditions. This results in a total energy usage for continual domain adaptation and AP50 that are comparable to the results of \texttt{w/o DGE}.

As the thresholds increase (i.e., 0.001 to 0.005 for DGTA and 0.01 to 0.02 for mBDD), there is a slight drop in AP50 (i.e., approximately 0.5\% for DGTA and 0.7\% for mBDD). However, \texttt{with DGE} can significantly reduce the total energy usage required for continual domain adaptation, by approximately 58\% for both DGTA and mBDD.

When the threshold is set at a substantially higher value (i.e., 0.04 for DGTA and 0.07 for mBDD), domain adaptation is not applied, resulting in zero total energy usage for continual domain adaptation. However, this also causes a considerable decrease in AP50, with approximately 4\% for DGTA and 6\% for mBDD, compared to the results of \texttt{w/o DGE}.

\vspace{0.5em}
\noindent\textbf{Comparison between \texttt{with DGE} and \texttt{w/o DGE}} \,\,\, We repeat each experiment 5 times, then report mean and standard deviation of AP50 and AR50 over 5 runs. The domain gap threshold is fixed to $0.02$ for all datasets. 

It is unsurprising that \texttt{SDA} and \texttt{UDA} consistently outperform \texttt{NoDA} by a significant margin in datasets DGTA, KITTI, sBDD, and mBDD (see Tables~\ref{tab:DGTA_results}, ~\ref{tab:kitti},~\ref{tab:sbdd} and~\ref{tab:mbdd}).

\begin{figure}
    \begin{tikzpicture}
        \begin{axis}[ybar,
                width=.5\textwidth,
                height=.3\textwidth,
                ymin=0,
                ymax=21000,        
                ylabel={Energy (kWh)},
                xtick=data,
                xticklabels = {
                   SDA\\w/o DGE,
                    SDA\\with DGE,
                    UDA\\w/o DGE,
                    UDA\\with DGE,
                },
                xticklabel style={font=\scriptsize, inner sep=0.2pt, align=center},
                major x tick style = {opacity=0},
                minor x tick num = 1,
                minor tick length=0ex,
                enlarge x limits={abs=0.5},
                bar width=2.7mm
                ]
        
        \addplot[draw=black,fill=myred] 
            table[x index=0,y index=1] \dgta; 
        \addplot[draw=black,fill=mygreen] 
            table[x index=0,y index=1] \sbdd; 
        \addplot[draw=black,fill=mybluelight] 
            table[x index=0] \mbdd; 
        \addplot[draw=black,fill=mylightyellow] 
            table[x index=0] \kitti; 
        \legend{DGTA, sBDD, mBDD, KITTI}
        \end{axis}
    \end{tikzpicture}
    \caption{Total energy usage for continual domain adaptation. Note that we only plot the mean of total energy usage as its standard deviation is insignificant.}
    \label{fig:energy_usage}
\end{figure}
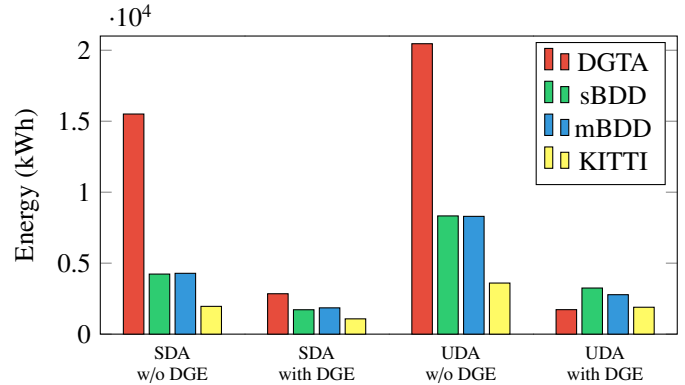

According to the results presented in Table~\ref{tab:DGTA_results} of the DGTA dataset, using \texttt{SDA} as the domain adaptation method,  \texttt{SDA w/o DGE} yields better overall performance than \texttt{SDA with DGE}, with an improvement of 1.1\% in AP50's mean and 0.9\% in AR50's mean. However, it is worth noting that the total energy consumption of \texttt{SDA w/o DGE} is approximately 81.6\% higher than that of \texttt{SDA with DGE}, as shown in Fig.~\ref{fig:energy_usage}. Similarly, if \texttt{UDA} is used as the domain adaptation method, Table~\ref{tab:DGTA_results} shows that the performance of \texttt{UDA w/o DGE} is comparable to that of  \texttt{UDA with DGE} in terms of AP50 and AR50. However, Fig.~\ref{fig:energy_usage} reveals that the energy consumption of \texttt{UDA with DGE}~is 91.6\% lower than that of \texttt{UDA w/o DGE}.

Regarding the KITTI dataset, Table~\ref{tab:kitti} reveals that \texttt{SDA w/o DGE} outperforms \texttt{SDA with DGE} by about 2\% in AP50's mean and 1.1\% in AR50's mean. Nonetheless, it is important to note that, as reported in Fig.~\ref{fig:energy_usage}, the total energy consumption of \texttt{SDA w/o DGE} is significantly higher, approximately 45\% more than \texttt{SDA with DGE}. Likewise, as indicated in Table~\ref{tab:kitti}, both \texttt{UDA w/o DGE} and \texttt{UDA with DGE} exhibit a similar level of accuracy. However, \texttt{UDA with DGE} achieves considerable energy savings, reducing total energy consumption by about 39\% in comparison to \texttt{UDA w/o DGE}.

When \texttt{SDA} is used as the domain adaptation method in the sBDD dataset, the performance of \texttt{SDA w/o DGE} is better than that of \texttt{SDA with DGE}, with an average improvement of 0.8\% in AP50 and 0.4\% in AR50, as reported in the Table~\ref{tab:sbdd}. However, it is important to note that the total energy consumption of \texttt{SDA w/o DGE} is 59.4\% higher than that of \texttt{SDA with DGE} (see Fig.~\ref{fig:energy_usage}). Also, Table~\ref{tab:sbdd} demonstrates that the AP50 and AR50 of \texttt{UDA with DGE} are similar to those of \texttt{UDA w/o DGE}. However, the comparison of the total energy consumption in Fig.~\ref{fig:energy_usage} reveals that \texttt{UDA with DGE} consumes significantly less energy than \texttt{UDA w/o DGE}.

The mBDD dataset analysis in Table~\ref{tab:mbdd} reveals that \texttt{SDA w/o DGE} marginally outperforms \texttt{SDA with DGE} by only 0.9\% in AP50's mean and 0.5\% in AR50's mean, which is statistically insignificant. However, as shown in Fig.~\ref{fig:energy_usage}, \texttt{SDA with DGE} offers significant energy savings, reducing total energy consumption by approximately 56.8\% compared to \texttt{SDA w/o DGE}. In addition, Table~\ref{tab:mbdd} shows that there is little difference in the performance of \texttt{UDA with DGE} and \texttt{UDA w/o DGE}, as measured by the AP50 and AR50 metrics. However, Fig.~\ref{fig:energy_usage} highlights the substantial energy savings achieved by \texttt{UDA with DGE}, reducing total energy usage by 66.5\% compared to \texttt{UDA w/o DGE}.

It is important to note that in the sBDD and mBDD datasets, the domain gap challenge presents a significant obstacle to achieving accuracy improvements through \texttt{UDA}. However, our paper's main focus is on determining ``when to adapt'' rather than ``how to adapt''. We believe that identifying the most appropriate times for adaptation can help overcome the domain gap challenge in future research.

\pgfplotstableread{
0 15506 
1 2846
2 20459
3 1725
}\dgta

\pgfplotstableread{
0 4231
1 1719
2 8328
3 3248
}\sbdd

\pgfplotstableread{
0 4285
1 1852
2 8297
3 2779
}\mbdd

\pgfplotstableread{
0 1957
1 1077
2 3599
3 1895
}\kitti

\section{Discussion and conclusion}

\subsection{Discussion} \label{sec:discussion}

\noindent\textbf{Selection of domain gap threshold} \,\,\, Effective application of DGE in continual domain adaptation involves a manual process of threshold selection, which is primarily due to two underlying factors: \emph{i)} The range of values for the metrics MMD, DSS, and SWD differs and is dependent on the dataset. Specifically, as shown in Fig.~\ref{fig:correlation}, for dataset DGTA, $\text{MMD} \in \left[ 0, 5.10^6 \right] $,  $\text{DSS} \in \left[ 0, 0.025 \right] $, and $\text{SWD} \in \left[0, 0.4 \right]$. However, for dataset sBDD or mBDD, $\text{MMD} \in \left[ 0, 10^6 \right] $,  $\text{DSS} \in \left[0, 0.07 \right]$, and $\text{SWD} \in \left[0, 0.2 \right]$. \emph{ii)} The impact of distinct thresholds on varying datasets is inconsistent. This is demonstrated in Fig.~\ref{fig:diff_thr}, where the relationship between total energy consumption for continual domain adaptation, the threshold for domain gap, and detection accuracy varies between two datasets, DGTA and mBD.

\vspace{0.5em}
\noindent\textbf{Catastrophic forgetting} \,\,\, Domain adaptation can lead to catastrophic forgetting, a phenomenon where the model's parameters are fine-tuned to fit the target domain, but at the cost of losing its knowledge of the source domain, resulting in poor performance on the source domain~\citep{mccloskey1989catastrophic}. To mitigate catastrophic forgetting, our paper examines the ideal scenario where all training data can be stored, although this is impractical in real-world applications. To address this issue, popular practical approaches such as replay-based, regularisation-based, and parameter isolation-based methods have been proposed (see~\citet{de2021continual} for a detailed survey). In light of these approaches, developing a DGE method that can be effectively combined with these methods is a promising direction for future research.

\vspace{0.5em}
\noindent\textbf{Test-time adaptation} \,\,\, Test-time adaptation (TTA) aims to adapt the model during the testing phase, when source data is inaccessible due to privacy concerns~\citep{tent, cotta, ttn}. As our work assumes the presence of source data, making it nontrivial to apply to TTA, developing a ``when to adapt'' method for TTA represents a promising avenue for further investigation.

\subsection{Conclusion}
This paper examines the use of maximum mean discrepancy, distance of second-order statistics, and sliced Wasserstein distance to evaluate the domain gap in object detection. The findings from experiments on synthetic and real-world datasets indicate a correlation between domain gap and detection accuracy. The paper then applies domain gap to the continual domain adaptation of RetinaNet, resulting in a significant reduction in the overall cost  of adaptation.

\section*{Acknowledgments}
This research was supported by the Australian Research Council grant LP200200881.
\bibliographystyle{model2-names}
\bibliography{bib}

\end{document}